\documentclass{article} 

\usepackage[utf8]{inputenc} 
\usepackage[T1]{fontenc}    
\usepackage{float}
\usepackage{url}            
\usepackage{booktabs}       
\usepackage{amsfonts}       
\usepackage{nicefrac}       

\usepackage{graphicx}
\usepackage{caption}
\usepackage{subcaption}
\DeclareCaptionType{noticebox}
\usepackage{amsmath,soul}
\usepackage{amsthm}
\usepackage{amssymb}
\usepackage{color}
\usepackage{paralist}
\usepackage{ifthen}
\usepackage{array}
\usepackage{epstopdf}
\usepackage{epsfig}
\usepackage{enumerate}
\usepackage{footmisc}
\usepackage{amsbsy}
\usepackage{bm}
\usepackage{array,cases}
\usepackage{multirow}
\usepackage{times}
\usepackage{booktabs}
\usepackage{algorithm}
\usepackage{algorithmic}
\usepackage{comment}
\usepackage{tabularx}

\newboolean{longver}
\setboolean{longver}{true}

\newcommand{\set}[1]{\ensuremath{\mathcal #1}}

\newcommand{\separator}{
  \begin{center}
    \rule{\columnwidth}{0.3mm}
  \end{center}
}

\newcommand{\beq}{\begin{eqnarray*}}
\newcommand{\eeq}{\end{eqnarray*}}
\newcommand{\beqn}{\begin{eqnarray}}
\newcommand{\eeqn}{\end{eqnarray}}
\newcommand{\bemn}{\begin{multiline}}
\newcommand{\eemn}{\end{multiline}}

\newcommand{\grad}[1]{\nabla #1}

\newtheorem{lemma}{Lemma}[section]

\newtheorem{theorem}{Theorem}

\newcommand\defeq{\mathrel{:=}}

\newcommand\ie{{\em i.e.}}
\newcommand\eg{{\em e.g.}}

\definecolor{webmag}{rgb}{0.5,0,0.5}
\newcommand{\nop}[1]{}

\title{Adiabatic Persistent Contrastive Divergence Learning}

\author{Hyeryung Jang$^\dag$, Hyungwon Choi$^\dag$, Yung Yi$^\dag$, and
  Jinwoo Shin$^\dag$\thanks{$\dag$: Department of Electrical
    Engineering, KAIST, South Korea, e-mails: hrjang@lanada.kaist.ac.kr,
    hyungwon.choi@kaist.ac.kr, yiyung@kaist.edu,
    jinwoos@kaist.ac.kr. Address for Correspondence: Jinwoo Shin, KAIST
    291, Daehak-ro, Yuseong-gu, Daejeon, 305-701, South Korea.}}

\allowdisplaybreaks

\begin{document}
\maketitle

\begin{abstract}
  This paper studies the problem of parameter learning in probabilistic
  graphical models having latent variables, where the standard approach
  is the expectation maximization algorithm alternating expectation (E)
  and maximization (M) steps. However, both {E} and {M} steps are
  computationally intractable for high dimensional data, while the
  substitution of one step to a faster surrogate for combating against
  intractability can often cause failure in convergence. We propose a
  new learning algorithm which is computationally efficient and provably
  ensures convergence to a correct optimum from the multi-time-scale
  stochastic approximation theory. Its key idea is to run only a few
  cycles of Markov Chains (MC) in both {E} and {M} steps. Such an idea
  of running `incomplete' MC has been well studied only for {M} step in
  the literature, called {\em Contrastive Divergence} (CD)
  learning. While such known CD-based schemes find approximated
  solutions via the mean-field approach in {E} step, our proposed
  algorithm does exact ones via MC algorithms in both
  steps. Consequently, the former maximizes an approximation (or lower
  bound) of log-likelihood, while the latter does the actual
  one. Despite of the theoretical understandings, the proposed scheme
  might suffer from the slow mixing of MC in {E} step. To address the
  issue, we also propose a hybrid approach adapting both mean-field and
  MC approximations in {E} step, and it outperforms the bare mean-field
  CD schemes in our experiments on real-world datasets.
\end{abstract}

\section{Introduction}

Graphical model (GM) has been one of powerful paradigms for succinct
representations of joint probability distributions in various
scientific fields including information theory, statistical physics
and artificial intelligence. GM represents a joint distribution of
some random variables by a graph structured model where each vertex
corresponds to a random variable and each edge captures the
conditional dependence between random variables. We study the problem
of learning parameters in graphical models having latent (or hidden)
variables. To this end, a standard learning procedure is the
expectation maximization (EM) algorithm alternating expectation (E)
and maximization (M) steps, where both involve certain inference
tasks. However, they are computationally intractable for
high-dimensional data.

To address the issue, Hinton et al.\ \cite{welling2002new,
  tieleman08PCD} suggested the so-called (persistent and
non-persistent) Contrastive Divergence (CD) learning algorithms based
on the stochastic approximation and mean-field theories. They apply
the mean-field approach in E step, and run an incomplete Markov chain
(MC) only few cycles in M step, instead of running the chain until it
converges or mixes. Consequently, the persistent CD maximizes (a
variational lower bound of) the log-likelihood, and the non-persistent
CD minimizes the reconstruction error induced by a few cycles of MC.
The authors also have demonstrated their performances in deep GMs such
as Restricted Boltzmann Machine (RBM) \cite{smolensky87rbm} and Deep
Boltzmann Machine (DBM) \cite{hinton09DBM} for various applications,
\eg, image \cite{larochelle2008classification}, speech
\cite{dahl10RBM} and recommendation \cite{salakhut07RBM}. In
principle, they are applicable for any GMs with latent variables.

In this paper, we propose a new CD algorithm, called {{\em Adiabatic
    Persistent Contrastive Divergence} (APCD)}. The design principle
can be understood as a `probabilistic' analogue of the standard
adiabatic theorem \cite{BF28} in quantum mechanics which states that
if a system changes in a reversible manner at an infinitesimally small
rate, then it always remains in its ground state. It is
computationally efficient and provably ensures convergence to a
correct optimum of the log-likelihood. While the persistent mean-field
CD maximizes a variational lower bound of the log-likelihood, the
proposed algorithm does the actual log-likelihood directly. Our key
idea is conceptually simple: run the exact MC method, instead of the
mean-field approximation, in E step as well. Namely, APCD runs
incomplete MCs in both {E} and {M} steps simultaneously. We prove that
it converges to a local optimum (or stationary point) of the actual
log-likelihood under mild assumptions by extending a standard
stochastic approximation theory \cite{borkar08SA} to the one with {\em
  multi-time-scales}. Such guarantee is hard to obtain under the known
mean-field CD learning since it optimizes a biased log-likelihood due
to the mean-field errors.

Despite the theoretical understandings, APCD might perform worse in
practice than the mean-field CD schemes since E step might take long
time to converge, \ie, slow mixing. To address the issue, we also
design a hybrid scheme that utilizes both the mean-field and MC
advantages in E step. Our experiments on real-world image datasets
demonstrate that APCD outperforms the bare mean-field CD scheme under
deep GMs. To our best knowledge, APCD is the first efficient algorithm
with provable convergence to a correct (local) optimum of
log-likelihood for general GMs. We anticipate that applications of our
new technique will be of interest to various fields where GMs with
latent variables are used for statistical modeling, while some
existing deep learning methods, which often perform similarly as APCD
in our experiments, were designed for more special purposes without
theoretical justifications.

There have been theoretical efforts to understand the CD learning in
the literature \cite{delyon99SAEM, kuhn04SAEM, younes99convergence,
  yuille04CD,tieleman10CD} under the stochastic approximation theory.
However, they assume that either E or M step is computable exactly,
while APCD runs incomplete MCs in both steps
simultaneously. Consequently, analyzing APCD becomes much more
challenging since some change in one step to a biased direction can
steer the other step towards a wrong direction. The main contribution
of our work is to overcome this technical challenge by adopting the
multi-time-scale stochastic approximation theory. We adjust learning
rates so that one of E and M steps runs in a faster time-scale than
the other, leading both steps to have correct estimations of the
gradient of log-likelihood.

There have also been several efforts to accelerate the CD schemes via
alleviating the slow mixing issue in M step. One of the most popular
techniques to boost mixing is simulated tempering
\cite{marinari1992simulated} which has also been studied in deep GMs
recently \cite{desjardins2010tempered, salakhutdinov2009learning,
  cho2010parallel, salakhutdinov2010learning}. The idea is to run a MC
with a slowly decreasing temperature under the intuition that the
high-temperature distributions are more spread out and easier to
sample from than the target distribution. These techniques can also be
applicable to the proposed APCD and hence are orthogonal to our work.

\section{Preliminaries}\label{sec:pre}

\subsection{Graphical model} \label{sec:GM}

\noindent{\bf Exponential family.} The exponential family is of our
interest, defined as follows.  We first let $\phi = (\phi_\alpha: \alpha
\in \mathcal{I})$ be a collection of real-valued functions $\phi_\alpha:
\mathcal{X} \rightarrow \mathbb{R}$ called {\em potential functions (or
  simply potentials)} or {\em sufficient statistics}, where $\mathcal{X}
\subset \mathbb{R}^k$ is the set of configurations. We assume that
$\set{X}$ is a finite set and $\phi$ is bounded. For a given vector of
sufficient statistics $\phi$, let $\theta = (\theta_\alpha: \alpha \in
\mathcal{I})$ be an associated vector called {\em canonical} or {\em
  exponential} parameters. For each fixed $x \in \mathcal{X}$, we use
$\langle \theta, \phi(x) \rangle$ to denote the inner product of two
vectors $\theta$ and $\phi(x)$, \ie, $\langle \theta, \phi(x) \rangle =
\sum_{\alpha \in \mathcal{I}} \theta_\alpha \phi_\alpha(x)$.  Using this
notation, the {\em exponential family} associated with the set of
potentials $\phi$ consists of the following collection of density
functions:
\begin{eqnarray} \label{eq:expfam}
p_\theta(x) = \exp \{ \langle \theta, \phi(x) \rangle - A(\theta) \}, \qquad\mbox{where}\quad   A(\theta) = \log \sum_{x \in \mathcal{X}} \exp \langle \theta, \phi(x) \rangle.
\end{eqnarray}
Here, $\exp\{A(\theta)\}$ is the normalizing constant called {\em
  partition function}. For a fixed potential vector $\phi$, each
parameter vector $\theta$ indexes a particular member $p_\theta$ of the
family. The canonical parameters $\theta$ of interest belong to the set
$ \Theta \defeq \{ \theta \in \mathbb{R}^{|\set{I}|}| A(\theta) < +
\infty \}. $

We assume {\em regularity} and {\em minimality} for the exponential
family throughout this paper, \ie, $\Theta$ is an open set and potential
functions $(\phi_\alpha: \alpha \in \set{I})$ are linearly
independent. For any regular exponential family, one can obviously check
that $A(\cdot)$ is a smooth and convex function of $\theta$, implying that
$\Theta$ must be a convex set. The minimality condition ensures that
there exists a unique parameter vector $\theta$ associated with each
density in the family.

\noindent{\bf Mean parameter.} It turns out that any exponential family
also allows an alternative parameterization by so-called {\em mean
  parameter} $\mu = (\mu_\alpha:\alpha \in \set{I})$. For any given
density function $p_\theta$, the mean parameter $\mu$ associated with a
sufficient statistic $\phi$ is defined by the following expectation:
\begin{eqnarray} \label{eq:mean param}
\mu_\alpha = \mathbb{E}_{\theta}[\phi_\alpha(X)] = \sum_{x \in
  \mathcal{X}} \phi_\alpha(x)p_{\theta}(x),\qquad\mbox{for $\alpha \in \set{I},$}
\end{eqnarray}
where we also define $\set{M_{\phi}} \defeq \left\{ \mu \in
  \mathbb{R}^{|\set{I}|} ~\Big|~ \exists\, p_\theta \ \text{such that} \
  \mu = \sum_{x \in \mathcal{X}} \phi(x)p_\theta(x) \right\},$ \ie,
$\set{M}_\phi$ is the set of all realizable mean parameters associated
with the given sufficient statistics $\phi.$

The gradient of the log partition function $A(\theta)$ has the following
connection to mean parameters:
\begin{eqnarray}
  \label{eq:der_a_theta}
  \frac{\partial A(\theta)}{\partial \theta_\alpha} &=& \mathbb{E}_{\theta}[\phi_\alpha(X)],\qquad\mbox{for $\alpha \in \set{I},$}
\end{eqnarray}
and therefore $\grad A(\theta) = \mu$, \ie, the mean parameters of
$p_\theta$. This can be viewed as a forward mapping from $\theta$ to
$\mu$. One can easily check that $\grad A: \Theta \mapsto
\mathcal{M}^\circ$ for any regular and minimal exponential family is a
bijection mapping. Moreover, we denote by $\theta^*: \mathcal{M}^\circ
\mapsto \Theta$ the inverse map of $\grad A$, \ie, $\theta^*(\mu) \defeq
\grad A^{-1}(\theta)$, thus $\mu =
\mathbb{E}_{\theta^*(\mu)}[\phi(X)]$. The existence and the
differentiability of $\theta^*$ is a direct consequence of the implicit
function theorem.

\subsection{Expectation maximization}

\noindent{\bf Learning exponential family.} For a given potential vector
$\phi,$ the goal is to learn exponential parameters $\theta$ given $N$
observed data $\mathbf{x} \defeq \{ x^n: n=1,\dots,N \}$, for which the
popular {\em Maximum Likelihood Estimation} (MLE) is used by solving the
following optimization problem:
\begin{eqnarray*}
  \text{{\bf MLE:}} \quad \theta^* = \text{arg} \max_{\theta \in \Theta}
  l(\theta; \mathbf{x}), \quad \text{where} \quad l(\theta; \mathbf{x})\defeq \frac{1}{N} \sum_{n=1}^N \log p_\theta(x^n).
\end{eqnarray*}
When computing the optimal solution $\theta^*,$ the gradient of the log-likelihood $l(\theta; \mathbf{x})$ has the following form due to
\eqref{eq:expfam} and \eqref{eq:der_a_theta}:
\begin{eqnarray}
  \label{eq:MLE_grad}
\frac{\partial l(\theta;\mathbf{x})}{\partial \theta} = \hat{\mu}-
  \mathbb{E}_\theta[\phi(X)], \qquad \text{where} \quad \hat{\mu} \defeq \frac{1}{N} \sum_{n=1}^N \phi(x^n).
\end{eqnarray}
Here, $\hat{\mu}$ is called {\em empirical mean parameter}. In many
applications of graphical models, a configuration $x \in \mathbb{R}^k$
tends to have a high dimension, often partially observed, thus some
units (\ie, coordinates) of $x$ are hidden (or latent). Thus, we denote
by $x = (v, h), \ v \in \mathcal{X}^v, h \in \mathcal{X}^h$ the entire
configuration with visible $v$ and hidden $h$ configurations, where
$\mathcal{X}^v$ and $\mathcal{X}^h$ are the domains of visible and
hidden ones, respectively. Clearly, $\mathcal{X} = \mathcal{X}^v \times
\mathcal{X}^h.$ Then, the probability density function of the
exponential family can be rewritten as $p_\theta(v,h) = \exp \{ \langle
\theta, \phi(v,h) \rangle - A(\theta) \}$, and denote by $ p_\theta(v)$
the density of a visible configuration $v$ marginalized over hidden
units, \ie, $p_\theta(v)= \sum_{h \in \mathcal{X}^h} p_\theta(v,h).$ In
presence of hidden units, for $N$ visible data $\mathbf{v} = \{ v^n:
n=1,\dots,N \},$ one aims at still learning parameters using the maximum
likelihood principle on {\em marginal} log-likelihood $l(\theta;
\mathbf{v})$:
\begin{align}
  \label{eq:MMLE}
  \text{{\bf MMLE:}} \quad \theta^* = \text{arg} \max_{\theta \in
    \Theta}l(\theta; \mathbf{v}), \quad \text{where} \quad l(\theta;
  \mathbf{v}) = \frac{1}{N} \sum_{n=1}^N \log
  p_\theta(v^n). 
\end{align}

\noindent{\bf Expectation Maximization.} A popular approach solving {\bf
  MMLE} is the {\em Expectation Maximization} (EM) algorithm. Consider a
distribution $q = \{ q^n(h): n=1,\dots,N \}$ over hidden units of each
visible data. Using Jensen's inequality, a lower bound of $l(\theta;
\mathbf{v})$ is given by:
   \begin{align}
     \label{eq:log_lbnd}
     l(\theta; \mathbf{v}) 
     &= \frac{1}{N} \sum_{n=1}^N \log \sum_{h \in \mathcal{X}^h} q^n(h) \frac{p_\theta(v^n,h)}{q^n(h)}
         \geq \frac{1}{N} \sum_{n=1}^N \sum_{h \in \mathcal{X}^h} q^n(h) \log \frac{p_\theta(v^n,h)}{q^n(h)} \cr
     &= \mathcal{F}(q,\theta) \defeq \frac{1}{N} \sum_{n=1}^N \Big( \sum_{h \in \mathcal{X}^h} q^n(h) \log p_\theta(v^n,h) + H(q^n) \Big),
   \end{align}
   where $H(q) = - \sum_{h \in \mathcal{X}^h} q(h) \log q(h)$ is the
   entropy of $q$. The EM algorithm, consisting of {E} and {M} steps,
   alternates between maximizing the lower bound
   $\mathcal{F}(q,\theta)$ with respect to $q$ and $\theta$,
   respectively, holding the other fixed: at each $t$-th iteration,
\begin{eqnarray*}
  \text{{\bf E step:}} \quad q_{(t+1)} = \text{arg} \max_{q} \mathcal{F}(q, \theta_{(t)}) \quad \quad
  \text{{\bf M step:}} \quad \theta_{(t+1)} = \text{arg} \max_{\theta} \mathcal{F}(q_{(t+1)}, \theta).
\end{eqnarray*}
E step reduces to inferring the probability of hidden units for each
given observed data, and it is well known that for exponential family
\eqref{eq:expfam}, the exact bound holds when $q_{(t+1)}^n(h)=
p_{\theta_{(t)}}(h|v^n)$ for each visible data $v^n$. Then, we compute
the expectation of $\phi(v^n,H)$, denoted by $\hat{\mu}^n_{(t+1)},$
where the random variable $H$ for a hidden configuration has the density
$p_{\theta_{(t)}}(h|v^n)$, and derive the empirical mean parameter
$\hat{\mu}$ (as in \eqref{eq:MLE_grad}), which is used in {M} step, \ie,
\begin{eqnarray*}
\hat{\mu}^n_{(t+1)} \defeq \sum_{h \in \mathcal{X}^h} \phi(v^n,h) p_{\theta_{(t)}}(h|v^n).
\end{eqnarray*}
{M} step now becomes equal to finding the canonical parameter in
{\bf MLE} (\ie, \eqref{eq:MLE_grad}), which is due to the fact that the
entropy of $q$ does not depend on $\theta$ in \eqref{eq:log_lbnd}.

Both {E} and {M} steps are computationally intractable in
general. First, {E} step requires deducing probability distribution over
hidden units from given canonical parameters, and exact inference
requires exponential time with respect to the number of hidden units. A
similar computational issue also arises in {M} step. The main
contribution of this paper is to develop a computationally efficient
learning algorithm that provably converges to a stationary point or
local optimum of {\bf MMLE}.

\section{Adiabatic persistent contrastive divergence}
\label{sec:APCD}

Now we are ready to present our main results: an algorithm to learn
exponential parameters $\theta$ for a given $\mathbf{v}$ and a graph
structure, and the theoretical analysis of the algorithm's
convergence. We first describe the algorithm and then show its provable
convergence guarantee.

\subsection{Algorithm description}

\begin{algorithm}[ht!]
  \caption{Adiabatic Persistent Contrastive Divergence (APCD): At each
    iteration $t = 0, 1, \ldots, $}
  \label{alg:APCD}
  \begin{algorithmic} \vspace{0.05in} 
    \STATE \textbf{Input:} Visible data $\mathbf{v} = \{v^n:
    n=1,\dots,N\}$, \\ \hspace{1cm}   $M$:  the number of MCs,
    $\ell$: the number of MC transitions to obtain a MC sample.\\
    \STATE \textbf{Output:} Canonical parameter $\theta_{(t)}$. \\
    \STATE \textbf{Initialize:} Set $\theta_{(0)} \in \Theta$, and $\{\hat{h}_{(0)}^{n,m}, \hat{x}_{(0)}^m, \hat{\mu}^n_{(0)}: n=1,\dots, N,m=1,\dots,M\}$ arbitrarily. \\
    \STATE \vspace{0.05in} \hrule
    \vspace{0.05in}
    \STATE {\bf \em /* E step */} 
    \vspace{0.05in}
    \FOR{$n=1$ \TO $N$ }
    \FOR{$m=1$ \TO $M$}
    \STATE \underline{{\bf \em E.1.}} Obtain a random sample
    $\hat{h}_{(t+1)}^{n,m}$ given $\hat{h}_{(t)}^{n,m}$ by taking $\ell$
    transitions of a time-reversible transition matrix
    $K^E_{\theta_{(t)},v^n}$ with the invariant distribution
    $p_{\theta_{(t)}}(h|v^n).$ Formally, 
    $$    p_{\theta_{(t)}}(h|v^n) = p_{\theta_{(t)}}(h|v^n)
    K^E_{\theta_{(t)},v^n} \quad \text{and} \quad \Pr\left[\hat{h}_{(t+1)}^{n,m}=h \mid \hat{h}_{(t)}^{n,m}\right] = (K^E_{\theta_{(t)},v^n})^{\ell}\big(\hat{h}_{(t)}^{n,m},h\big).$$
    \ENDFOR \\
    \vspace{0.05in}
    \STATE \underline{{\bf \em E.2.}} Update per-data empirical mean parameter $\hat{\mu}_{(t+1)}^n$ with the step-size $a_{(t)}$: 
    {\footnotesize \begin{eqnarray} \label{eq:E}
        \hat{\mu}_{(t+1)}^n &=& \hat{\mu}_{(t)}^n + a_{(t)} \left(\frac{1}{M}\sum_{m=1}^M \phi \Big(v^n,
        \hat{h}_{(t+1)}^{n,m}\Big) - \hat{\mu}_{(t)}^n \right).
      \end{eqnarray}}
    \ENDFOR
    \vspace{0.05in}
    \STATE \underline{{\bf \em E.3.}} Update the empirical mean
    parameter as: 
    $\hat{\mu}_{(t+1)} = \frac{1}{N}\sum_{n=1}^N \hat{\mu}_{(t+1)}^n.$
    
    \vspace{0.1in}
    \STATE {\bf \em /* M step */}
    \vspace{0.05in}
    \FOR{$m=1$ \TO $M$ }
    \STATE \underline{\bf \em M.1.} Obtain a random sample $\hat{x}_{(t+1)}^{m}$ given $\hat{x}_{(t)}^{m}$ by running $\ell$ transitions of a time-reversible transition matrix
    $K^M_{\theta_{(t)}}$ with the invariant distribution
    $p_{\theta_{(t)}}(x)$. Formally (and similarly to \underline{\bf \em E.1.}),
    $$p_{\theta_{(t)}}(x) = p_{\theta_{(t)}}(x)
    K^M_{\theta_{(t)}} \quad  \text{and} \quad  \Pr\left[\hat{x}_{(t+1)}^{m}=x \mid \hat{x}_{(t)}^{m}\right] =
    (K^M_{\theta_{(t)}})^\ell \big(\hat{x}_{(t)}^{m},x\big). \quad $$
    \ENDFOR \\
    \vspace{0.03in}
    \STATE \underline{\bf \em M.2.} Update the canonical parameter with the step-size $b(t) $ as:
    \begin{equation} \label{eq:M} 
      \theta_{(t+1)} =  \theta_{(t)} + b_{(t)} \bigg( \hat{\mu}_{(t+1)} - \frac{1}{M}\sum_{m=1}^M
      \phi(\hat{x}_{(t+1)}^m) \bigg).
    \end{equation}
  \end{algorithmic}
\end{algorithm}

The formal description of the proposed algorithm is given in {\bf
  Algorithm \ref{alg:APCD}}, which is a randomized version of the EM
algorithm using suitable step-size functions $a,b: \mathbb{Z}_{\geq 0}
\rightarrow \mathbb{R}^{+}$.  It can be interpreted as stochastic
approximation procedures based on MC method with different time-scales
in {E} and {M} steps, which we call {\bf \em Adiabatic Persistent
  Contrastive Divergence} (APCD) algorithm. It first obtains random
samples of hidden nodes and updates the empirical mean parameter vector
$\hat{\mu}$ in {E} step. Then, it obtains random samples of the entire
nodes and updates the parameter $\theta$ in {M} step. We provide more
details in what follows.

\smallskip
\noindent{\bf E step.} In ({\bf \em E.1.}) of $t$-th iteration, for each
visible data $v^n$, we first construct $M$ number of Markov chains with
transition matrix $K^E_{\theta_{(t)},v^n}$, each of which has
$p_{\theta_{(t)}}(h|v^n)$ as a stationary (or invariant) distribution,
and obtain a sample $\hat{h}_{(t+1)}^{n,m}$ at $m$-th MC by taking
$\ell$ transitions from the previous configuration
$\hat{h}^{n,m}_{(t)}$, \eg, Gibbs sampling. Each MC sampling is done by
clamping the values of visible nodes to each visible data $v^n$, and
running $\ell$ transitions of $K^E_{\theta_{(t)},v^n}$. Then, in ({\bf
  \em E.2.}), the algorithm updates per-data empirical mean parameter,
denoted as $\hat{\mu}_{(t+1)}^n$, by {\em (i)} sample-averaging of
corresponding sufficient statistics, and {\em (ii)} moving average with
step-size constant $a_{(t)}$. In ({\bf \em E.3.}), the empirical mean
parameter $\hat{\mu}_{(t+1)}$ is computed by taking its average over
data.

\smallskip
\noindent{\bf M step.} In {M} step of $t$-th iteration, the algorithm
computes stochastic gradient to update the canonical parameter, where
the gradient is \eqref{eq:MLE_grad} with empirical mean parameter of
$\hat{\mu}_{(t+1)}$. Similarly to ({\bf \em E.1.}), in ({\bf \em M.1.}),
we construct $M$ number of MCs with transition matrix
$K^M_{\theta_{(t)}}$, each of which has $p_{\theta_{(t)}}(x)$ as a
stationary distribution, and obtain a sample $\hat{x}_{(t+1)}^{m}$ at
$m$-th MC by taking $\ell$ transitions from the previous configuration
$\hat{x}_{(t)}^{m}$. Note that this step is independent of visible data
$\mathbf{v}$. Then, in ({\bf \em M.2.}), canonical parameters are
updated by {\em (i)} sample-averaging of entire sufficient statistic
vectors, and {\em (ii)} using it in running the gradient-ascent method
with step-size constant $b_{(t)}$.

\subsection{Convergence analysis}
\label{sec:convergence}

We now state the following convergence property of the proposed APCD
algorithm.

\begin{theorem}
  \label{thm:EM} 
  Choose positive step-size functions $a_{(t)}, b_{(t)}>0$ satisfying
    \begin{eqnarray}
      \label{eq:step-size}
      \sum_t a_{(t)} = \sum_t b_{(t)} = \infty, \quad \sum_t (a_{(t)}^2 + b_{(t)}^2) \leq \infty, \quad \frac{a_{(t)}}{b_{(t)}} \rightarrow \text{ either } 0 \text{ or } \infty.
    \end{eqnarray}
    Assume that $\{\theta_{(t)}\}$ and $\{\hat{\mu}_{(t)}\}$ remain
    bounded, almost surely. Then, under APCD,
    $\theta_{(t)}$ almost surely converges to a stationary point of {\bf
      MMLE}, \ie, a stationary point of $l(\theta; \mathbf{v})$ in
    \eqref{eq:MMLE}.
\end{theorem}
We remark that the above theorem does not guarantee that APCD converges
to a local optimum, \ie, it might stuck at a saddle point. However, APCD
is a stochastic gradient ascent algorithm and unlikely converges to a
saddle point.

The proof of Theorem~\ref{thm:EM} is given in the supplementary material
due to the space limitation, where we provide its proof sketch in this
section. A simple insight is that the conditions of the step-size
functions in Theorem \ref{thm:EM} require that MCs in one step should
run in a faster time-scale than those in the other step. When E step
takes a faster time-scale, the faster loop evaluates the averaged
empirical distribution for a given observed visible data $\mathbf{v}$
and the slowly-varying parameter value, and the slower loop in M step
finds the {\bf MLE} parameter which fits the averaged empirical
distribution evaluated at the faster loop. The examples of step-size
functions satisfying \eqref{eq:step-size} include $a_{(t)} = 1/t,
b_{(t)} = 1 / (1+t \log t)$, or $a_{(t)} = 1/ t^{2/3}, b_{(t)} = 1/t$.

{\bf Proof sketch.} Our main proof strategy is to follow the stochastic
approximation procedure with multi-time-scales whose limiting behavior
is understood by ordinary differential equations (ODE)
\cite{borkar08SA}. To this end, define a map from the discrete times of
{E} and {M} step to the real ones: $\alpha(t) = \sum_{i=0}^{t-1}
a_{(i)}$ and $\beta(t) = \sum_{i=0}^{t-1} b_{(i)}$, respectively. We
also denote by $\{ \hat{\mu}_\alpha(\tau), {\theta}_\alpha(\tau): \tau
\in \mathbb R_+ \}$ and $\{ \hat{\mu}_\beta(\tau), {\theta}_\beta(\tau):
\tau\in \mathbb R_+ \}$ the corresponding continuous-time linear
interpolations of $\{\hat{\mu}_{(t)},\theta_{(t)}: t\in \mathbb Z_{\geq
  0}\}$ for each time-scale $\alpha$ and $\beta$, respectively. The
convergence analysis of APCD is complicated in the sense that both E and
M steps include random Markov processes with different time-scales, \ie,
MC transitions are controlled by the current canonical parameter. Here
we provide a proof sketch when E step has a faster time-scale, \ie,
${b_{(t)}}/{a_{(t)}} \rightarrow 0$. The proof when E step has a slower
time-scale follows similar arguments.

As the first step, under the faster time-scale $\alpha$, the updates of
the slower loop in M step will be seen {\em quasi-static} for
sufficiently large $\tau$. This is because the dynamics of the slower
loop is rewritten as
\begin{eqnarray*}
\theta_{(t+1)} = \theta_{(t)} + a_{(t)} \cdot \left[
  \frac{b_{(t)}}{a_{(t)}} \left( \hat{\mu}_{(t)} - \frac{1}{M}
    \sum_{m=1}^M \phi(\hat{x}_{(t+1)}^m) \right) \right],
\end{eqnarray*}
and its limiting ODE system for $\alpha$ is $\dot{\theta}(\tau) =
0$. Then, the dynamics of E step $\hat{\mu}_\alpha(\tau)$ tracks the
following ODE system of $\mu(\tau)$, where the behavior of the slower
loop (M step) is fixed to a quasi-static value, say $\theta$, and the MC
in E step is seen equilibrated with its invariant distribution
$p_\theta(h|\mathbf{v})$\footnote{For simplicity, we use $f(\mathbf{v})$
  to denote the average over observed data, \ie, $\frac{1}{N}
  \sum_{n=1}^N f(v^n)$.}:
\begin{eqnarray} \label{eq:E-fast-ode}
\dot{\mu}(\tau) = \sum_{h \in \mathcal{X}^h} \phi(\mathbf{v},h)
p_\theta(h|\mathbf{v}) - \mu(\tau).
\end{eqnarray}
We analyze asymptotic convergence of the faster loop by showing that the
ODE \eqref{eq:E-fast-ode} has a unique fixed point
$\hat{\mu}^*(\theta;\mathbf{v}) \defeq \sum_{h \in \mathcal{X}^h}
\phi(\mathbf{v},h) p_\theta(h|\mathbf{v})$, \ie, the expectation of
empirical mean parameter over the distribution $p_\theta(h|\mathbf{v})$,
thus we have almost surely $\hat{\mu}_{(t)} \rightarrow
\hat{\mu}^*(\theta; \mathbf{v})$.

As the second step, under the slower time-scale $\beta$, the behavior of
the faster loop $\hat{\mu}_\beta(\tau)$ would appear to be equilibrated
for the current quasi-static $\theta_\beta(\tau)$, \ie,
$\hat{\mu}_\beta(\tau) \approx
\hat{\mu}^*(\theta_\beta(\tau);\mathbf{v})$. Then, the dynamics of M
step $\theta_\beta(\tau)$ tracks the following ODE system of
$\theta(\tau)$, where the behavior of the faster loop and MC in M step
are equilibrated to $\hat{\mu}^*(\theta(\tau);\mathbf{v})$ and
$p_{\theta(\tau)}(x)$, respectively:
\begin{eqnarray} \label{eq:M-slow-ode}
  \dot{\theta}(\tau) = \hat{\mu}^*(\theta(\tau);\mathbf{v}) - \sum_{x \in \mathcal{X}} \phi(x)p_{\theta(\tau)}(x).
\end{eqnarray}
We show that the ODE \eqref{eq:M-slow-ode} has a Lyapunov function
$V(\theta) = -l(\theta; \hat{\mu}^*(\theta;\mathbf{v}))$, specifically a
negative log-likelihood with empirical mean parameter
$\hat{\mu}^*(\theta; \mathbf{v})$, which is indeed a marginal
log-likelihood $l(\theta;\mathbf{v})$. Then, from the known results on
Lyapunov function of stochastic approximation procedure, we have almost
surely $\theta_{(t)} \rightarrow \{ \theta: \partial_\theta V(\theta) =
0 \}$. Combining these results, we derive that under APCD,
$\theta_{(t)}$ almost surely converges to a stationary point of {\bf
  MMLE}.

\section{Experimental results} \label{sec:experiment}

We compare the APCD algorithm with the popular mean-field persistent
contrastive divergence (MFPCD) algorithm \cite{welling2002new,
  tieleman08PCD,hinton09DBM}, where they differ only in E step.  We
consider the pairwise binary graphical model over graph $G=(V,E)$: 
$$ p_\theta(x) \propto \exp \left(\sum_{i \in V} \theta_i x_i + \sum_{(i,j)
    \in E} \theta_{ij} x_ix_j \right)$$ for $x \in \{0,1\}^{|V|}.$ We
first consider grid models with randomly selected hidden units for
synthetic datasets in Section~\ref{sec:grid}, and then consider Deep
Boltzmann Machine (DBM) \cite{hinton09DBM} with two hidden layers for
real-world image datasets, MNIST, OCR letters, Frey Face, and Toronto
Face (TF) in Section~\ref{sec:deep}.

\noindent {\bf Basic setup.} We commonly use the popular Gibbs sampler
for the time-reversible transition matrix in both E and M steps. In M
step of both MFPCD and APCD, we use $\ell = 10$, $M = 100$ as in
\cite{hinton09DBM}. In E step of APCD, we use $\ell = 100$, $M = 1$
for the update of the per-data empirical mean parameter, while we run
$30$ mean-field iterations in E step of MFPCD. In addition, we choose
step-sizes which decreases linearly at every epoch but with different
speed for E and M step, as Theorem~\ref{thm:EM}
suggested. Specifically, we use the popular choice of $b_{(t)}$ in M
step, which is well studied in MFPCD, and choose $a_{(t)}$ decaying
$10$ times faster. Then, APCD is slower than MFPCD by roughly $3$
times in E step, and in overall, $2 \sim 3$ times slower per each
epoch in our simulation.

\subsection{Shallow models on synthetic datasets} \label{sec:grid}

We report our experimental results of APCD for the two dimensional grid
graph $G$ of size $|V|=30 \times 30$, where $[\theta_i]_{i \in V}$ is
set to random values in range $[-3.0,3.0]$ and $[\theta_{ij}]_{(i,j) \in
  E}$ is set to random Gaussian values with mean $0$ and variance
$0.5$. Then, under the random choice of parameters, we generate $2,000$
synthetic samples for training and another $2,000$ samples for test by
running the Gibbs sampler with $50,000$ iterations for each. We consider
two models, each with a different portion of hidden variables: among
$30\times 30$ variables, we randomly select $50\%$ and $20\%$ of them as
hidden ones. We train the synthetic dataset by AFCD and MFPCD equally
for $300$ epochs. For M step, the initial learning rate (\ie, step-size)
is set to be $0.001$ and decay gradually to $0.0001$. That for E step
starts from $1$ and decreases to $0.05$, so that we run E step at a
faster time-scale.  Finally, we generate $2,000$ samples from each
trained model and use {\em Parzen window} density estimation
\cite{breuleux2011quickly} to measure the average log-likelihood of the
test data. The $\sigma$ parameter in the Parzen method is cross
validated, where we use $20\%$ of the training set as validation.

\begin{figure}[!t]
\begin{subfigure}[b]{0.45\textwidth}
\centering
\includegraphics[width=\columnwidth]{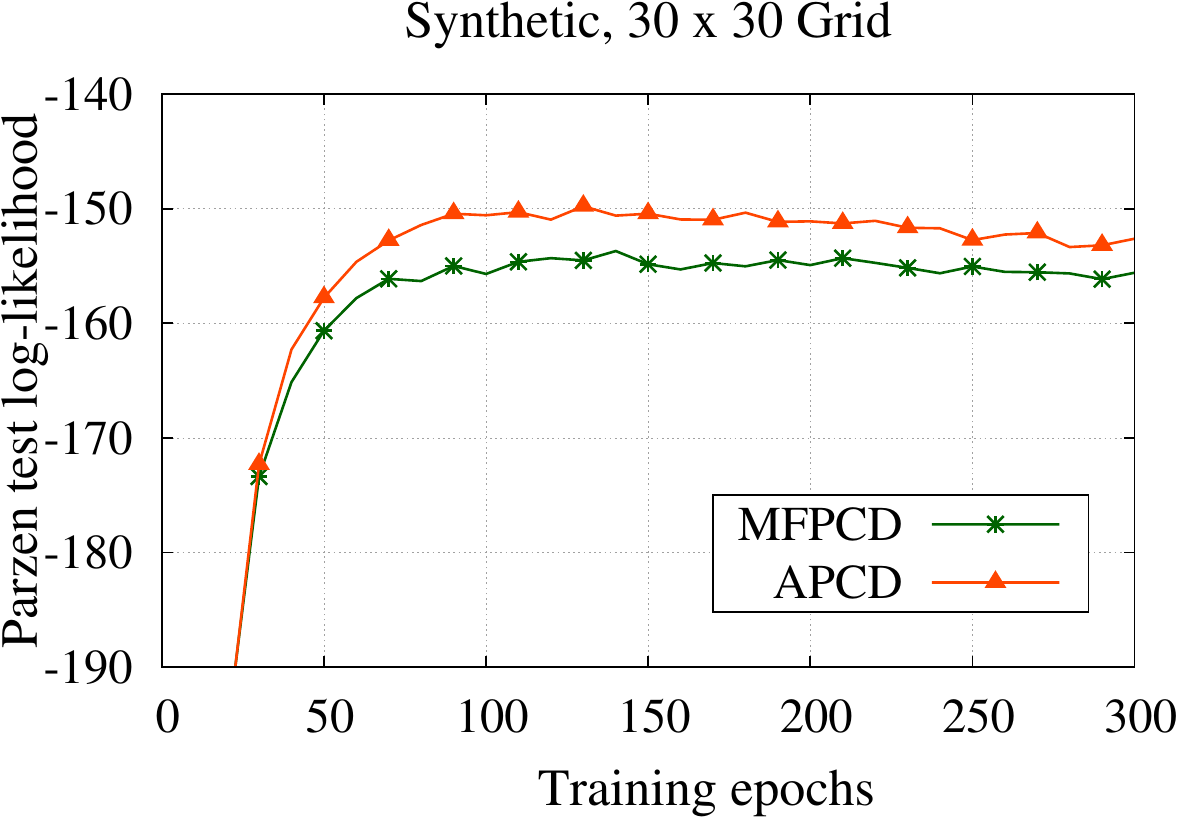}
\caption{Grid model on synthetic data: Average test log-likelihood of
  MFPCD and APCD trained model on synthetic dataset in every $10$
  epochs.}
\label{fig:parzen_grid}
\end{subfigure}
\hfill
\begin{subfigure}[b]{0.45\textwidth}
\centering
\includegraphics[width=\columnwidth]{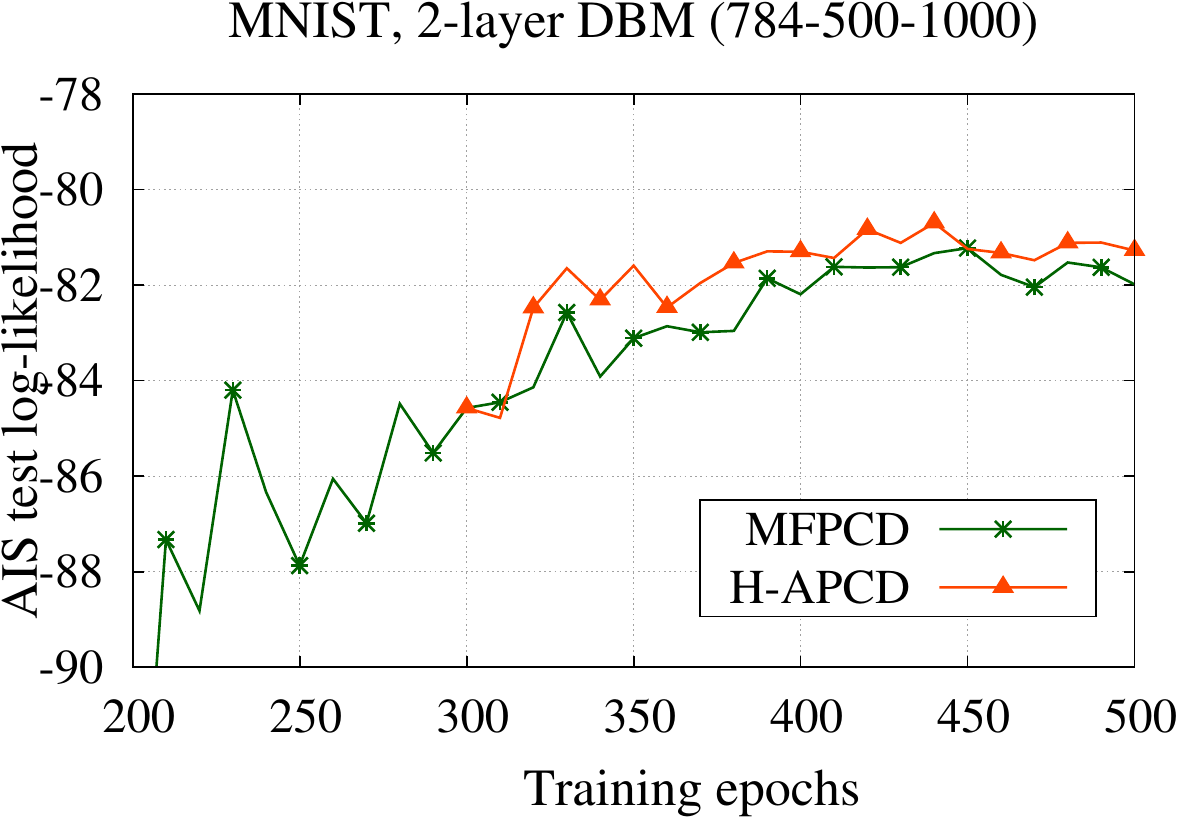}
\caption{DBM on MNIST: Average test log-likelihood of MFPCD and H-APCD
  trained model on MNIST dataset in every $10$ epochs.}
\label{fig:logprob}
\end{subfigure}
\caption{Generative performances in grid and deep models per training epoch.}
\end{figure}

\noindent{\bf Generative performance.} For the first $50\%$-hidden
model, the Parzen log-likelihood estimates for APCD and MFPCD are
$-149.79 \pm 0.35$ and $-153.70 \pm 0.33$ ($\pm$ indicates the standard
error of the mean computed across examples). On the other hand, the
Parzen measure obtained on the training set is $-148.90 \pm 0.35$, \ie,
close to that of APCD. As reported in Figure~\ref{fig:parzen_grid}, APCD
starts to outperform MFPCD after $50$ training epochs.  The Parzen
estimates for the second $20\%$-hidden model trained by APCD and MFPCD
are $-256.10 \pm 0.41$ and $-260.16 \pm 0.41$, respectively. In this
case, the reference measure on the training set is $-254.26 \pm
0.42$. These results demonstrate that APCD provides major improvements
over MFPCD in these synthetic settings.

\subsection{Deep models on real-world datasets} \label{sec:deep}

We now report our experimental results of APCD for Deep Boltzmann
Machine (DBM). We train two-hidden-layer DBM on the following datasets:
MNIST, OCR letters, Frey Face, and Toronto Face (TF). MNIST dataset
contains $60,000$ training images and $10,000$ test images of
handwritten digits. We use 10,000 images from the training set for
validation. OCR letters dataset consists of images of $26$ English
characters. The dataset is split into $32,152$ training, $10,000$
validation, and $10,000$ test samples.  Frey Face and TF datasets are
both real-valued grey-scale images of human faces. While Frey is
relatively small, containing $1,965$ images in total, TF contains almost
$100,000$ images. For Frey, we use $1,800$ images for training and
$10\%$ of the training set as validation. For TF, we follow the splits
provided by the dataset.

In order to train a $2$-layer DBM with MFPCD, we follow the same
hyperparameter settings as described in \cite{hinton09DBM, salakhut10BM}
for MNIST and OCR. For Frey and TF, the model architecture used in our
experiments is $560$-$200$-$400$ and $2304$-$500$-$1000,$
respectively. Pretraining of DBM is performed for 100 epochs over the
training set\footnote{For Frey and TF, we use Gaussian-Binary Restricted
  Boltzmann Machines (GBRBM) for pretraining as described in
  \cite{nair09RBM}.} and the global training is done for $500$, $200$,
$200$, and $400$ epochs for MNIST, OCR, Frey and TF, with minibatch size
of $100$. For {M} step, the initial learning rate is set to be $0.005$
and decay gradually to $0.0001$ as training progresses.  For the DBM
experiments, we use Annealed Importance Sampling (AIS)
\cite{neal2001annealed} and Parzen window density estimation to measure
the average log-likelihood of the test data.

In APCD, we basically follow the same hyperparameters of MFPCD in {M}
step. The {E} step learning rate starts from a large value close to $1$,
and decreases slowly to $0.05$. We also design the following practical
hybrid training scheme for DBM, which we call Hybrid APCD (H-APCD).  We
first train DBM via MFCD in the first halfway in the whole training
steps. Then, in the second half, we take the weighted sum of the
probabilities computed from APCD and MFPCD, where the ratio of such
fusion gradually changes not to favor MFPCD as training progresses.  The
reason why we take such a hybrid approach of APCD and MFPCD is due to
our observation that estimations of E steps in APCD are initially bad in
large DBMs due to the mixing issue.  We note that for grid graphs in the
previous section, such a hybrid training is not necessary since the
models are relatively small.

\noindent{\bf Generative performance.} In Table \ref{tbl:logprob}, we
compare the average test log-likelihood of MFPCD and H-APCD along with
other previous works. For MNIST and OCR, we first run AIS $100$ times to
estimate the model partition function.  Then, we run $100$ AIS runs
separately for each test sample to estimate the test log-likelihood.  We
randomly sample $1,000$ images\footnote{ We measure the true
  log-likelihood instead of its variational bound (although it takes
  much more time) for fair comparison between APCD and MFPCD.} from the
test set to measure the average test log-likelihood. For Frey and TF, we
only report Parzen estimates since calculating the log-likelihood using
AIS with Gaussian DBM is not straightforward. "True" in
Table~\ref{tbl:logprob} is computed by running Parzen estimates on
10,000 random samples from the training set. We generate $10,000$
samples from each trained model for Parzen estimates.

\begin{table}[t!]
\begin{center}
\vskip -0.1in
\hspace{-0.2cm}
\scriptsize
\caption{Generative performances of MFPCD and H-APCD. (a) Test
  log-likelihood of $1,000$ random samples from the test set measured by
  running separate $100$ AIS runs per data. (b) Parzen window-based
  log-likelihood estimates conducted as in
  \cite{goodfellow2014generative}.} \label{tbl:logprob} \vspace{0.25cm}
\begin{subtable}{0.4\columnwidth}\centering
    \begin{tabular}{lcc } 
      \toprule
      & MNIST & OCR \\ \midrule
      {\bf MFPCD} & $-84.31$ & $-31.13$ \\ \midrule 
      {\bf H-APCD} & $-83.93$ & $-29.59$  \\ 
     \bottomrule\\
    \end{tabular}
    \vspace{-0.1in}
\caption{}\label{tab:1a}
\vspace{0.15cm}
\end{subtable}
\hfill
\hspace{-0.6cm}
\begin{subtable}{0.7\columnwidth}\centering
\begin{tabular}{lcccc} 
  \toprule
  & MNIST & OCR & Frey & TF \\ \midrule
  {\bf DBN\cite{hinton2006fast}} & $138 \pm 2$ &.&.& $1909 \pm 66$ \\ \midrule 
  {\bf Stackted CAE\cite{bengio2013better}} & $121 \pm 1.6$ &.&.& $2110 \pm 50$ \\ \midrule 
  {\bf Deep GSN\cite{bengio2014deep}} & $214 \pm 1.1$ &.&.& $1890 \pm 29$ \\ \midrule 
  {\bf GAN\cite{goodfellow2014generative}} & $225 \pm 2$ &.&.& $2057 \pm 26$ \\ \midrule 
  
  {\bf MFPCD} & $239.48 \pm 1.7$ & $-52.68 \pm 0.3$ & $659 \pm 11$ & $1939 \pm 28$ \\ \midrule 
  {\bf H-APCD} & $239.24 \pm 2.1$ & $-52.66 \pm 0.4$ & $684 \pm 11$ & $1985 \pm 42$ \\ \midrule
  {\bf True} & $244 \pm 1.9$ & $-27 \pm 0.4$ & $931 \pm 18$ & $2119 \pm 23$\\
 \bottomrule
\end{tabular}

\caption{}\label{tab:1b}
\vspace{-0.3cm}
\end{subtable}
\vspace{-0.55cm}
\end{center}
\end{table}

For MNIST and OCR, H-APCD exceeds MFPCD in terms of test log-likelihood
of $1,000$ test samples measured by AIS, and performs similarly on
Parzen estimates. Figure \ref{fig:logprob} shows the average test
log-likelihood of MFPCD and H-APCD trained model in every $10$ epochs on
MNIST dataset. The log-likelihood of H-APCD exceeds that of MFPCD after
a small amount of training steps and the gap continues to exist until
the end of the training. For Frey and TF, H-APCD performs well with a
larger margin. The result is comparable to other previous works as well
as the true Parzen estimates.


\section{Conclusion}
In this paper, we propose a new efficient algorithm for parameter
learning in graphical models with latent variables.  Unlike other known
similar methods, it provably converges to a correct optimum. We believe
that our techniques based on the multi-time-scale stochastic
approximation theory should be of broader interest for designing and
analyzing similar algorithms.


\begin{small}
\bibliographystyle{unsrt}
\bibliography{ref}

\begin{thebibliography}{10}

\bibitem{welling2002new}
M.~Welling and G.~E. Hinton.
\newblock A new learning algorithm for mean field boltzmann machines.
\newblock In {\em Artificial Neural Networks—ICANN 2002}, pages 351--357.
  Springer, 2002.

\bibitem{tieleman08PCD}
T.~Tieleman.
\newblock Training restricted boltzmann machines using approximations to the
  likelihood gradient.
\newblock In {\em Proceedings of the International Conference on Machine
  Learning}, pages 1064--1071, 2008.

\bibitem{smolensky87rbm}
P.~Smolensky.
\newblock Information processing in dynamical systems: Foundations of harmony
  theory.
\newblock In D.~E. Rumelhart, J.~L. McClelland, et~al., editors, {\em Parallel
  Distributed Processing}, pages 194--281. MIT Press, 1987.

\bibitem{hinton09DBM}
R.~Salakhutdinov and G.~E. Hinton.
\newblock Deep boltzmann machines.
\newblock In {\em Proceedings of the International Conference on Artificial
  Intelligence and Statistics}, pages 448--455, 2009.

\bibitem{larochelle2008classification}
H.~Larochelle and Y.~Bengio.
\newblock Classification using discriminative restricted boltzmann machines.
\newblock In {\em Proceedings of the International Conference on Machine
  Learning}, pages 536--543, 2008.

\bibitem{dahl10RBM}
G.~Dahl, A.~R. Mohamed, and G.~E. Hinton.
\newblock Phone recognition with the mean-covariance restricted boltzmann
  machine.
\newblock In {\em Proceedings of the Advances in Neural Information Processing
  Systems}, pages 469--477, 2010.

\bibitem{salakhut07RBM}
R.~Salakhutdinov, A.~Mnih, and G.~E. Hinton.
\newblock Restricted boltzmann machines for collaborative filtering.
\newblock In {\em Proceedings of the International Conference on Machine
  Learning}, pages 791--798, 2007.

\bibitem{BF28}
M.~Born and V.~A. Fock.
\newblock Beweis des adiabatensatzes.
\newblock {\em Zeitschrift fur Physik a Hadrons and Nuclei}, 51(3-4):165–180,
  1928.

\bibitem{borkar08SA}
V.~S. Borkar, editor.
\newblock {\em Stochastic Approximation: A Dynamical Systems Viewpoint}.
\newblock Cambridge University Press, 2008.

\bibitem{delyon99SAEM}
B.~Delyon, M.~Lavielle, and E.~Moulines.
\newblock Convergence of a stochastic approximation version of the {EM}
  algorithm.
\newblock {\em Annals of Statistics}, 27(1):94--128, 1999.

\bibitem{kuhn04SAEM}
E.~Kuhn and M.~Lavielle.
\newblock Coupling a stochastic approximation version of {EM} with an mcmc
  procedure.
\newblock {\em ESAIM: Probability and Statistics}, 8:115--131, 2004.

\bibitem{younes99convergence}
L.~Younes.
\newblock On the convergence of markovian stochastic algorithms with rapidly
  decreasing ergodicity rates.
\newblock {\em Stochastics: An International Journal of Probability and
  Stochastic Processes}, 65(3-4):177--228, 1999.

\bibitem{yuille04CD}
A.~Yuille.
\newblock The convergence of contrastive divergences.
\newblock In {\em Proceedings of the Advances in Neural Information Processing
  Systems}, 2004.

\bibitem{tieleman10CD}
I.~Sutskever and T.~Tieleman.
\newblock On the convergence properties of contrastive divergence.
\newblock In {\em Proceedings of the International Conference on Artificial
  Intelligence and Statistics}, pages 789--795, 2010.

\bibitem{marinari1992simulated}
E.~Marinari and G.~Parisi.
\newblock Simulated tempering: a new {M}onte {C}arlo scheme.
\newblock {\em EPL (Europhysics Letters)}, 19(6):451, 1992.

\bibitem{desjardins2010tempered}
G.~Desjardins, A.~Courville, Y.~Bengio, P.~Vincent, and O.~Delalleau.
\newblock Tempered {M}arkov chain {M}onte {C}arlo for training of restricted
  {B}oltzmann machines.
\newblock In {\em Proceedings of the International Conference on Artificial
  Intelligence and Statistics}, pages 145--152, 2010.

\bibitem{salakhutdinov2009learning}
R.~Salakhutdinov.
\newblock Learning in {M}arkov random fields using tempered transitions.
\newblock In {\em Proceedings of the Advances in Neural Information Processing
  Systems}, pages 1598--1606, 2009.

\bibitem{cho2010parallel}
K.~Cho, T.~Raiko, and A.~Ilin.
\newblock Parallel tempering is efficient for learning restricted {B}oltzmann
  machines.
\newblock In {\em Proceedings of the International Joint Conference on Neural
  Networks}, pages 1--8, 2010.

\bibitem{salakhutdinov2010learning}
R.~Salakhutdinov.
\newblock Learning deep {B}oltzmann machines using adaptive {MCMC}.
\newblock In {\em Proceedings of the International Conference on Machine
  Learning}, pages 943--950, 2010.

\bibitem{breuleux2011quickly}
O.~Breuleux, Y.~Bengio, and P.~Vincent.
\newblock Quickly generating representative samples from an rbm-derived
  process.
\newblock {\em Neural Computation}, 23(8):2058--2073, 2011.

\bibitem{salakhut10BM}
R.~Salakhutdinov and H.~Larochelle.
\newblock Efficient learning of deep botlzmann machines.
\newblock In {\em Proceedings of the International Conference on Artificial
  Intelligence and Statistics}, pages 693--700, 2010.

\bibitem{nair09RBM}
V.~Nair and G.~E. Hinton.
\newblock Implicit mixtures of restricted boltzmann machines.
\newblock In {\em Proceedings of the Advances in Neural Information Processing
  Systems}, pages 1145--1152, 2009.

\bibitem{neal2001annealed}
R.~M. Neal.
\newblock Annealed importance sampling.
\newblock {\em Statistics and Computing}, 11(2):125--139, 2001.

\bibitem{goodfellow2014generative}
I.~Goodfellow, J.~Pouget-Abadie, M.~Mirza, B.~Xu, D.~Warde-Farley, S.~Ozair,
  A.~Courville, and Y.~Bengio.
\newblock Generative adversarial nets.
\newblock In {\em Proceedings of the Advances in Neural Information Processing
  Systems}, pages 2672--2680, 2014.

\bibitem{hinton2006fast}
G.~E. Hinton, S.~Osindero, and Y.~Teh.
\newblock A fast learning algorithm for deep belief nets.
\newblock {\em Neural computation}, 18(7):1527--1554, 2006.

\bibitem{bengio2013better}
Y.~Bengio, G.~Mesnil, Y.~Dauphin, and S.~Rifai.
\newblock Better mixing via deep representations.
\newblock In {\em Proceedings of the International Conference on Machine
  Learning}, pages 552--560, 2013.

\bibitem{bengio2014deep}
Y.~Bengio, E.~Laufer, G.~Alain, and J.~Yosinski.
\newblock Deep generative stochastic networks trainable by backprop.
\newblock In {\em Proceedings of the International Conference on Machine
  Learning}, pages 226--234, 2014.

\bibitem{proutiere10RA}
A.~Proutiere, Y.~Yi, T.~Lan, and M.~Chiang.
\newblock Resource allocation over network dynamics without timescale
  separation.
\newblock In {\em Proceedings of the 29th Conference on Information
  Communications (INFOCOM 2010)}, pages 406--410. IEEE Press, 2010.

\end{thebibliography}
\end{small}

\appendix
\section*{Appendix: Proof of Theorem \ref{thm:EM}} \label{sec:proof}

The convergence analysis of the Adiabatic Persistent Contrastive
Divergence (APCD) is on the strength of multi-time-scale stochastic
approximation theory. As we mentioned in Section~\ref{sec:APCD}, our
algorithm is interpreted as a stochastic approximation procedure with
controlled Markov processes. In this supplementary material, we first
provide the convergence analysis of a general stochastic approximation
procedure (\ie, with single time-scale) with a controlled Markov process
in Section~\ref{sec:SA-CMN}, where an ordinary differential equation
(ODE) is usefully utilized to study the limiting behavior of the system
states. Then, in Section~\ref{sec:APCD-proof}, we non-trivially extend
this framework to the APCD algorithm, where the step-size conditions in
Theorem~\ref{thm:EM}, controlling the speed of the two dynamics in E and
M steps respectively, are the key to the convergence proof.

\section{Preliminary: stochastic approximation with controlled Markov
  process} \label{sec:SA-CMN}

Consider a discrete-time stochastic process $\{x_{(t)}: t \in
\mathbb{Z}_{\geq 0}\}$ with the following form:
\begin{eqnarray} 
  \label{eq:SA-cont} 
  x_{(t+1)} = x_{(t)} + a_{(t)} \cdot v(x_{(t)},Y_{(t+1)}), \quad
  \forall t \in \mathbb{Z}_{\geq 0},
\end{eqnarray}
where $x_{(t)} \in \mathbb{R}^L$ is the system state at the iteration
$t$, $a_{(t)}$ corresponds to the step-size, and the system has a Markov
process 
taking values in finite space $\mathcal{Z}$ with control process
$x_{(t)}$, \ie, with a controlled transition kernel $K^{x_{(t)}}$. At
iteration $t$, the Markov process generates $M$ realizations $\{
\hat{z}_{(t+1)}^m: m=1,\cdots,M \}$ from the succession of $\ell$ cycles
of the transition kernel $K^{x_{(t)}}(\hat{z}_{(t)}^m, \cdot)$, \ie, 
\begin{eqnarray*}
  \Pr\left[\hat{z}_{(t+1)}^{m}=z \mid \hat{z}_{(t)}^{m}\right] =
  (K^{x_{(t)}})^\ell \big(\hat{z}_{(t)}^{m},z\big).
\end{eqnarray*}
Then, the observation $Y_{(t+1)}$ is a function of the random samples
$\{ \hat{z}_{(t+1)}^m\}$ as:
\begin{eqnarray*}
Y_{(t+1)} = \frac{1}{M} \sum_{m=1}^M f(\hat{z}_{(t+1)}^m).
\end{eqnarray*}
This is often called {\em stochastic approximation with controlled
  Markov process} \cite{borkar08SA}. We shall assume that if $x_{(t)} =
x, \forall t$ for a fixed $x \in \mathbb{R}^L$, the Markov process is
irreducible and ergodic with unique invariant distribution $\pi^x$, and
let $\zeta^x(\mathrm{d}y)$ denote by the stationary distribution of
$\frac{1}{M} \sum_{m=1}^M f(\hat{z}^m)$, where $\{\hat{z}^m\}$ are drawn
from the Markov process controlled by $x$. In addition, we assume that:


\begin{compactenum}[\bf (C1)]
\item For any $x \in \mathbb{R}^L$, $x \mapsto K^x$ is continuous and $x
  \mapsto \pi^x$ is Lipschitz continuous.
\smallskip
\item The function $f:\mathcal{Z} \mapsto \mathbb{R}^K$ is a bounded,
  and $v(x,Y): \mathbb{R}^{L+K} \mapsto \mathbb{R}^L$ is a bounded
  Lipschitz continuous in $x$ and uniformly over $Y$.
\smallskip
\item Almost surely, $\{x_{(t)}\}$ remains bounded.
\smallskip
\item $\{a_{(t)}\}$ is a decreasing sequence of positive number such
  that $\sum_t a_{(t)} = \infty$ and $\sum_t a_{(t)}^2 < \infty$.
\end{compactenum}
Note that there exist many dynamical MC-based procedures, \eg, Gibbs
sampler, Metropolis-Hasting algorithm and variants, which provide
property of {\bf (C1)}, and {\bf (C3)} can be imposed by projecting the
process to a bounded subset of $\mathbb{R}^L$. The example choices of a
step-size (or learning rate) function include $a_{(t)} = \frac{1}{t},
\frac{1}{t^{2/3}}, \frac{1}{1+t \log t}$.

Now, define $\alpha(t) = \sum_{i=0}^{t-1} a_{(i)}$. We take a
continuous-time pairwise linear interpolation of the system state $\{
x_{(t)}: t \in \mathbb{Z}_{\geq 0}\}$ under the time-scale $\alpha$ in
the following way: define $\{ x_\alpha(\tau): \tau \in \mathbb{R}_+ \}$
as: $\forall t \in \mathbb{Z}_{\geq 0}$, for all $\tau \in
[\alpha(t),\alpha(t+1))$,
\begin{eqnarray} \label{eq:int}
  x_\alpha(\tau) = x_{(t)} + (x_{(t+1)}-x_{(t)}) \times \frac{\tau - \alpha(t)}{\alpha(t+1) - \alpha(t)}. 
\end{eqnarray}

\paragraph{Remark 1.} Intuitively, for a decreasing step-size $a_{(t)}$,
the interpolated continuous trajectory $x_\alpha(\tau)$ is an
accelerated version of the original trajectory $x_{(t)}$. Note that as
the decreasing speed of $a_{(t)}$ becomes faster, \ie, $a_{(t)}
\rightarrow 0$ at a faster rate, the stochastic process
\eqref{eq:SA-cont} moves on a slower time-scale.




Now, the following theorem provides the convergence analysis of the
stochastic approximation procedure with controlled Markov process
\eqref{eq:SA-cont}. 

\medskip
\begin{theorem} [Theorem 1 of \cite{proutiere10RA}, Corollary 8 in
  Chapter 6.3 of \cite{borkar08SA}] 
\label{thm:SA-CMN}
Suppose that assumptions {\bf (C1)-(C4)} hold. Let $T > 0$, and denote
by $\tilde{x}^s(\cdot)$ the solution on $[s,s+T]$ of the following
ordinary differential equation (ODE):
\begin{eqnarray} \label{eq:SA-CMN-ode}
\dot{x}(\tau) = \int_y v(x(\tau),y) \cdot \zeta^{x(\tau)}(\mathrm{d}y), \quad
\text{with} \quad \tilde{x}^s(s) = x_\alpha(s),
\end{eqnarray}
Then, we have almost surely,
\begin{eqnarray*}
\lim_{s \rightarrow \infty} \sup_{\tau \in [s,s+T]} \| x_\alpha(\tau) - \tilde{x}^s(\tau) \| = 0.
\end{eqnarray*}
Moreover, $x_{(t)}$ converge a.s. to an internally chain transitive
invariant set of the ODE \eqref{eq:SA-CMN-ode}.
\end{theorem}

Note that since the Markov process is irreducible and ergodic, and $f$
is continuous and bounded, we have almost surely,
\begin{eqnarray*}
  \int_y v(x,y) \zeta^x(\mathrm{d}y) = \sum_{z \in \mathcal{Z}} v(x,f(z))\pi^{x}(z).
\end{eqnarray*}
Therefore, the ODE \eqref{eq:SA-CMN-ode} becomes the following simpler
form, which will be used later in the proof of Theorem~\ref{thm:EM}:
\begin{eqnarray} \label{eq:SA-CMN-ode2}
\dot{x}(\tau) &=& \sum_{z \in \mathcal{Z}} v(x(\tau),f(z)) \pi^{x(\tau)}(z).
\end{eqnarray}





\paragraph{Remark 2.} 
We comment that there exists a slight difference between the model of
this section and that of in \cite{proutiere10RA} and \cite{borkar08SA}:
The controlled Markov process is a discrete-time one in our setup,
whereas it is a continuous-time one in \cite{borkar08SA,proutiere10RA},
requiring just a simple modification of the proof.



Theorem~\ref{thm:SA-CMN} states that as time evolves, the dynamics of
the underlying Markov process is {\em averaged} due to the decreasing
step-size, thus ``almost reaching the stationary regime''. Thus, it
suffices to see how the ODE \eqref{eq:SA-CMN-ode2} behaves. In
particular, when the ODE \eqref{eq:SA-CMN-ode2} has the unique fixed
stable equilibrium point $x^*$, we have almost surely: $x_{(t)} \to x^*$
as $t \to \infty$. If the ODE \eqref{eq:SA-CMN-ode2} has a {\em
  Lyapunov} function, then every internally chain transitive invariant
lies in the Lyapunov set, and thus the process \eqref{eq:SA-cont}
converges to the largest internally chain transitive invariant set.


\section{Proof of Theorem \ref{thm:EM}} \label{sec:APCD-proof}

We now prove the convergence of Algorithm~\ref{alg:APCD} by showing that
it is a multi-time-scale stochastic approximation procedure with
controlled Markov process. We will use some result that works for
general multi-time-scale stochastic approximation procedure
\cite{borkar08SA}. To that end, we rewrite E and M step of APCD into
following form:
\begin{eqnarray} \label{eq:SAP-E}
  \text{{\bf E step:}} \quad \hat{\mu}_{(t+1)} &=& \hat{\mu}_{(t)} +
  a_{(t)} \cdot g\bigg( \hat{\mu}_{(t)}, \theta_{(t)}, H_{(t+1)} \bigg),
\end{eqnarray}
\begin{eqnarray} \label{eq:SAP-M}
  \text{{\bf M step:}} \quad \theta_{(t+1)} &=& \theta_{(t)} + b_{(t)}
  \cdot u \bigg( \hat{\mu}_{(t)}, \theta_{(t)}, X_{(t+1)}  \bigg),
\end{eqnarray}
where 
\begin{eqnarray*}
g(\hat{\mu},\theta,H) = H - \hat{\mu}, \quad u(\hat{\mu},\theta,X) =
\hat{\mu} - X, 
\end{eqnarray*}
with
\begin{eqnarray*}
H_{(t+1)} = \frac{1}{N}\sum_{n=1}^N H_{(t+1)}^n, \quad H_{(t+1)}^n =
\frac{1}{M} \sum_{m=1}^M \phi(v^n, \hat{h}_{(t+1)}^{n,m}), \quad
X_{(t+1)} = \frac{1}{M} \sum_{m=1}^M \phi(\hat{x}_{(t+1)}^m).
\end{eqnarray*}
Note that for each visible data $v^n$, $\{\hat{h}_{(t+1)}^{n,m}: m =
1,\cdots,M\}$ are samples generated from the successive $\ell$ cycles of
transition matrix $K_{\theta_{(t)},v^n}^E$ in E step, and
$\{\hat{x}_{(t+1)}^m: m=1,\cdots,M\}$ are samples generated from the
successive $\ell$ cycles of transition matrix $K_{\theta_{(t)}}^M$ in M
step. One can easily check that $u(\cdot), g(\cdot)$ are bounded
Lipschitz continuous for exponential family with bounded sufficient
statistics $\phi$.

We now analyze the coupled stochastic approximation procedures
\eqref{eq:SAP-E} and \eqref{eq:SAP-M}, under the following two
``time-scales'' with different speed: (i) $\alpha(t) = \sum_{i=0}^{t-1}
a_{(i)}$ and (ii) $\beta(t) = \sum_{i=0}^{t-1} b_{(i)}$. We denote by
$\{\hat{\mu}_\alpha(\tau),\theta_\alpha(\tau): \tau \in \mathbb{R}_+ \}$
and $\{\hat{\mu}_\beta(\tau),\theta_\beta(\tau): \tau \in \mathbb{R}_+
\}$ the corresponding continuous-time interpolations of
$\{\hat{\mu}_{(t)}, \theta_{(t)}: t \in \mathbb{Z}_{\geq 0}\}$ according
to \eqref{eq:int} for time-scales $\alpha$ and $\beta$,
respectively. The condition $\frac{a_{(t)}}{b_{(t)}} \to 0$ or
$\frac{a_{(t)}}{b_{(t)}} \to \infty$ in Theorem~\ref{thm:EM} implies
that the decreasing speed of two steps are different: \ie, one step
should run in a faster time-scale than the other step. If
$\frac{a_{(t)}}{b_{(t)}} \to 0$, E step moves at a slower time-scale
than M step, and if $\frac{a_{(t)}}{b_{(t)}} \to \infty$, \ie,
equivalently $\frac{b_{(t)}}{a_{(t)}} \to 0$, E step moves at a faster
time-scale than M step.


\medskip
\subsection{Case 1: $\frac{a_{(t)}}{b_{(t)}} \to 0$}

From the hypothesis $\frac{a_{(t)}}{b_{(t)}} \to 0$, we can first prove
the following two properties:
\begin{compactenum}[\bf P1.]
\item For all $T>0$, almost surely\footnote{Here, $|| \cdot ||$ corresponds to the $L_2$-norm.},
\begin{eqnarray}
\label{eq:the1}
\lim_{s \to \infty}\sup_{\tau \in [s,s+T]}\|\hat{\mu}_\beta(\tau) - \hat{\mu}_\beta(s) \| = 0.
\end{eqnarray}
\item  Almost surely, 
\begin{eqnarray*}
  \lim_{\tau \to \infty} \| \theta_\alpha(\tau) - \theta^*(\hat{\mu}_\alpha(\tau)) \| = 0.
\end{eqnarray*}
\end{compactenum}

{\bf P1} states that $\hat{\mu}_\beta(\tau)$ almost behaves like a
constant after a sufficient number of iterations. This is due to the
fact that $\hat{\mu}_{(t)}$ is updated by the step-size $a_{(t)},$ but
$\hat{\mu}_\beta(\tau)$ is the trajectory made by the faster time-scale
of $b_{(t)}$. More formally, by rewriting \eqref{eq:SAP-E}, we have:
\begin{eqnarray*}
  \hat{\mu}_{(t+1)} &=& \hat{\mu}_{(t)} + b_{(t)} \cdot \bigg[ \frac{a_{(t)}}{b_{(t)}} g\bigg( \hat{\mu}_{(t)}, \theta_{(t)}, H_{(t+1)} \bigg) \bigg], \\
\end{eqnarray*}
and thus it is obvious that its limiting ODE is $\dot{\mu}(\tau) = 0.$
Then, the property {\bf P1} immediately holds. {\bf P2} implies that
$\theta_\alpha(\tau)$ is asymptotically close to a unique fixed point
$\theta^*(\hat{\mu}_\alpha(\tau))$, a {\bf MLE} parameter in
\eqref{eq:MLE_grad} for a given empirical mean parameter
$\hat{\mu}_\alpha(\tau)$. Note that for a regular and minimal
exponential family, the map $\theta^*(\cdot)$ is a bijection mapping,
see Section~\ref{sec:GM}. In the rest of the proof, we first show {\bf
  P2} in {\bf \em Step 1}, and then in {\bf \em Step 2}, we complete the
proof of \underline{\bf Case 1} using {\bf P1} and {\bf P2}.




\medskip

{\bf \em Step 1:} \underline{\em Understanding the asymptotic behavior
  of the system at the faster time-scale $\beta$.}

We now introduce $\{\theta_\beta^s(\tau): \tau \in \mathbb{R}_+\}$,
which interpolates $\{\theta_{(t)}^s: t \in \mathbb{Z}_{\geq 0} \}$
(similarly to \eqref{eq:int}), where $\{\theta_{(t)}^s: t \in
\mathbb{Z}_{\geq 0}\}$ is constructed such that with $s \in \mathbb{R}$,
$\theta_{(t)}^s = \theta_{(t)}$ for $s \geq \beta(t)$, and
\begin{eqnarray} \label{eq:the2}
 \theta^s_{(t+1)} = \theta^s_{(t)} + b_{(t)} \cdot u\bigg( \hat{\mu}_\beta(s), \theta^s_{(t)}, X_{(t+1)} \bigg),
\end{eqnarray}
for $s < \beta(t)$. Note that \eqref{eq:the2} is different to
\eqref{eq:SAP-M} in that $\hat{\mu}_{(t)}$ is fixed to
$\hat{\mu}_\beta(s)$. Then, from the Lipschitz continuity of $u(\cdot)$,
we get that for all $T > 0$:
\begin{eqnarray} \label{eq:the3}
\lim_{s \to \infty}\sup_{\tau \in [s,s+T]}\|\theta_\beta^s(\tau) - \theta_\beta(\tau) \| = 0.
\end{eqnarray}


Now, we will compare $\theta_\beta^s(\tau)$ to the solution trajectory
of the following ODE, as in \eqref{eq:SA-CMN-ode2} of
Section~\ref{sec:SA-CMN}:
\begin{eqnarray} \label{eq:ODE-M1}
\dot{\theta}(\tau) = u^\dagger(\theta(\tau)) \defeq \sum_{x \in \mathcal{X}} u \bigg( \hat{\mu}_\beta(s), \theta(\tau), \phi(x) \bigg) p_{\theta(\tau)}(x).
\end{eqnarray}
To explain how our setup matches with that in Section~\ref{sec:SA-CMN},
let $\tilde{\theta}^s(\tau)$ be the solution on $[s,s+T]$ (for $T>0$) of
the ODE \eqref{eq:ODE-M1} with $\tilde{\theta}^s(s) =
\theta_\beta^s(s)$. It is clear that $\{\theta_{(t)}^s: t \in
\mathbb{Z}_{\geq 0}\}$ is a discrete-time stochastic process with
controlled Markov process considered in \eqref{eq:SA-cont} for $s \leq
\beta(t)$. We can verify that the assumptions {\bf (C1)-(C4)} are
satisfied. First, the MC sampler in M step $K_\theta^M$ induces an
ergodic Markov chain taking values in a finite set $\mathcal{X}$, which
satisfies the detailed balance property with respect to $p_\theta(x)$ in
\eqref{eq:expfam} for a fixed $\theta \in \Theta$, \eg, Gibbs
sampler. {\bf (C1)} is verified with our choice of $K_\theta^M$. {\bf
  (C2)} is verified since sufficient statistics $\phi$ is bounded. {\bf
  (C3)} is the assumption of Theorem~\ref{thm:EM}, where some techniques
for establishing this stability condition has been studied, see Chapter
3 of \cite{borkar08SA}. Finally, {\bf (C4)} is verified with our choice
of $b_{(t)}$. Then, we have that for all $T > 0$,
\begin{eqnarray*}
  \lim_{s \rightarrow \infty} \sup_{\tau \in [s,s+T]} \|
  \theta_\beta(\tau) - \tilde{\theta}^s(\tau) \| = 0, \quad \text{ a.s.}.
\end{eqnarray*}
This is a direct consequence of Theorem~\ref{thm:SA-CMN} and
\eqref{eq:the3}.

Now, for any regular and minimal exponential family, the ODE system
\eqref{eq:ODE-M1} has a unique fixed point $\theta^* =
\theta^*(\hat{\mu}_\beta(s))$, \ie, a {\bf MLE} parameter in
\eqref{eq:MLE_grad} for a given empirical mean parameter
$\hat{\mu}_\beta(s)$, due to the fact that
\begin{eqnarray*}
  u^\dagger(\theta(\tau)) = \hat{\mu}_\beta(s) -
  \mathbb{E}_{\theta(\tau)}[\phi(X)] = \grad l(\theta(\tau);
  \hat{\mu}_\beta(s)). 
\end{eqnarray*}

Then, we apply the following Lemma in Chapter 6.1 of \cite{borkar08SA}:
\begin{lemma} [Lemma 1 in Chapter 6.1 of \cite{borkar08SA}]
  \label{lem:step1}
  \begin{eqnarray*}
    (\hat{\mu}_{(t)}, \theta_{(t)}) \rightarrow \{ (\hat{\mu},
    \theta^*(\hat{\mu})): \hat{\mu} \in \mathcal{M}^\circ  \}, \quad \text{ a.s.}.
  \end{eqnarray*}
\end{lemma}
In other words, we have almost surely,
\begin{equation} \label{eq:fast-M}
\| \theta_{(t)} - \theta^*(\hat{\mu}_{(t)}) \| \rightarrow 0, 
\end{equation}
which in turn implies that under the slower time-scale $\alpha$, we also
have almost surely,
\begin{equation} \label{eq:step1-M}
\lim_{\tau \to \infty} \| \theta_\alpha(\tau) - \theta^*(\hat{\mu}_\alpha(\tau)) \|
= 0. 
\end{equation} 
This completes the proof of {\bf P2}.



\medskip

{\bf \em Step 2.} \underline{\em Understanding the asymptotic behavior
  of the system at the slower time-scale $\alpha$.}

We start by introducing $\{\hat{\mu}_\alpha^s(\tau); \tau \in
\mathbb{R}_+\}$ (similar to $\theta_\beta^s(\tau)$ in {\bf \em Step 1}),
which interpolates $\{\hat{\mu}_{(t)}^s: t \in \mathbb{Z}_{\geq 0}\}$,
where it is constructed such that with $s \in \mathbb{R}$,
$\hat{\mu}_{(t)}^s = \hat{\mu}_{(t)}$ for $s \geq \alpha(t)$, and 
\begin{eqnarray*} \label{eq:mu1}
 \hat{\mu}^s_{(t+1)} = \hat{\mu}^s_{(t)} + a_{(t)} \cdot g\bigg( \hat{\mu}^s_{(t)}, \theta^* (\hat{\mu}^s_{(t)}), H_{(t+1)} \bigg),
\end{eqnarray*}
for $s < \alpha(t)$. Note that \eqref{eq:mu1} is different to
\eqref{eq:SAP-E} in that $\theta_{(t)}$ is fixed to
$\theta^*(\hat{\mu}_{(t)}^s)$. Then, from \eqref{eq:step1-M} and the
fact that $g(\cdot)$ is Lipschitz continuous, it follows that for all $T
> 0$,
\begin{eqnarray} \label{eq:mu2}
\lim_{s \to \infty}\sup_{\tau \in [s,s+T]}\|\hat{\mu}_\alpha^s(\tau) - \hat{\mu}_\alpha(\tau) \| = 0. 
\end{eqnarray}




Now, we compare $\hat{\mu}_\alpha^s(\tau)$ to the solution trajectory of
the following ODE, as in Section~\ref{sec:SA-CMN}:
\begin{eqnarray} \label{eq:ODE-E1}
\dot{\mu}(\tau) = g^\dagger(\mu(\tau)) \defeq \sum_{h \in \mathcal{X}^h} g \bigg( \mu(\tau), \theta^* (\mu(\tau)), \phi(\mathbf{v},h) \bigg) p_{\theta^*(\mu(\tau))}(h|\mathbf{v}).
\end{eqnarray}
It is clear that the process $\{ \hat{\mu}_{(t)}^s: t \in
\mathbb{Z}_{\geq 0} \}$ is also a discrete-time stochastic process with
controlled Markov process considered in \eqref{eq:SA-cont} for $s \leq
\alpha(t)$. The assumptions {\bf (C1)-(C4)} are verified as follows:
The MC sampler in E step $K_{\theta,\mathbf{v}}^E$\footnote{For
  simplicity, we denote the set of $K_{\theta,v^n}^E$ for each visible
  data $v^n$ by $K_{\theta, \mathbf{v}}^E$.} induces an ergodic Markov
chain taking values in a finite set $\mathcal{X}^h$, which satisfies the
detailed balance property with respect to $p_\theta(h|\mathbf{v})$ for a
fixed $\theta \in \Theta$, \eg, Gibbs sampler. With our choice of
$K_{\theta, \mathbf{v}}^E$, since $\phi$ is bounded and
$\theta^*(\cdot)$ is continuous, we have $\hat{\mu} \mapsto
p_{\theta^*(\hat{\mu})}(h|\mathbf{v})$ and becomes Lipschitz continuous,
\ie, {\bf (C1)} is verified. {\bf (C2),(C3)} are verified since $\phi$
is bounded. Finally, {\bf (C4)} is verified with our choice of
$a_{(t)}$. Let denote by $\tilde{\mu}^s(\tau)$ be the solution on
$[s,s+T]$ (for any $T > 0$) of the ODE \eqref{eq:ODE-E1} with
$\tilde{\mu}^s(s) = \hat{\mu}_\alpha^s(s)$. Then, we now have that for
all $T > 0$, 
\begin{eqnarray*}
  \lim_{s \rightarrow \infty} \sup_{\tau \in [s,s+T]} \|
  \hat{\mu}_\alpha(\tau) - \tilde{\mu}^s(\tau) \| = 0, \quad \text{ a.s.},
\end{eqnarray*}
which is a direct consequence of Theorem~\ref{thm:SA-CMN} and
\eqref{eq:mu2}.

Next, we claim that there exists a {\em Lyapunov function} of the ODE
\eqref{eq:ODE-E1}. 

\begin{lemma}
\label{lem:SAEM-lyap} 
Assume that exponential family $(\theta,\phi)$ has regularity,
minimality and bounded sufficient statistics. Then, $V(\hat{\mu}) =
-l(\theta^*(\hat{\mu});\mathbf{v})$ is a Lyapunov function of the ODE
\eqref{eq:ODE-E1}: \ie, $V(\hat{\mu})$ is a function such that
  \begin{compactenum}[(a)]
  \item For all $\hat{\mu} \in \mathcal{M}$, $F(\hat{\mu}) \defeq
    \langle \partial_{\hat{\mu}}V(\hat{\mu}),g^{\dagger}(\hat{\mu})
    \rangle \leq 0$,
\smallskip
  \item $V( \{ \hat{\mu}: F(\hat{\mu}) = 0 \} )$ has an empty interior.
  \end{compactenum}
  Moreover,
  \begin{eqnarray*}
    \{ \hat{\mu}: F(\hat{\mu}) = 0 \} = \{ \hat{\mu}: \partial_{\hat{\mu}}V(\hat{\mu}) = 0 \} \quad\mbox{and}\quad
    \theta^*( \{ \hat{\mu}: F(\hat{\mu}) = 0 \} ) = \{ \theta \in
    \Theta: \partial_\theta l(\theta;\mathbf{v}) = 0 \}.
  \end{eqnarray*}
\end{lemma}

\paragraph{Remark 3.} Lemma~\ref{lem:SAEM-lyap} is an application of
Lemma 2 presented in \cite{delyon99SAEM}, which gives a result about
Lyapunov function of the ODE system of the form \eqref{eq:ODE-E1}. We
omit the formal proof of Lemma~\ref{lem:SAEM-lyap} since the assumptions
of Lemma 2 in \cite{delyon99SAEM} are shortly verified for a regular and
minimal exponential family with bounded sufficient statistics.


We now use the following result in \cite{borkar08SA} at our framework to
discuss the convergence guarantee of $\{ \hat{\mu}_{(t)}: t \in
\mathbb{Z}_{\geq 0} \}$ and the convergence point characterization.

\begin{lemma} [Corollary 3 in Chapter 2.2 of \cite{borkar08SA}]
  \label{cor:lyap} Under the assumption that $\{\hat{\mu}_{(t)}\}$
  remains bounded, $\{ \hat{\mu}_{(t)} \}$ almost surely converges to an
  internally chain transitive invariant set contained in $\{ \hat{\mu}
  \in \mathcal{M}^\circ: F(\hat{\mu}) = 0 \}$.
\end{lemma}

By Lemma~\ref{lem:SAEM-lyap} and Lemma~\ref{cor:lyap}, it follows that 
\begin{eqnarray} \label{eq:E-lyap}
   \hat{\mu}_{(t)} \rightarrow \{
   \hat{\mu}: \partial_{\hat{\mu}}V(\hat{\mu}) = 0 \}, \quad \text{ a.s.}.
 \end{eqnarray}
 This completes the proof of {\bf \em Step 2}.

\medskip

Combining {\bf \em Step 1} and {\bf \em Step 2}, \ie, from
\eqref{eq:fast-M} and \eqref{eq:E-lyap}, we complete the proof for
\underline{\bf Case 1} of Theorem \ref{thm:EM} that under APCD,
$\theta_{(t)}$ almost surely converges to a stationary point of {\bf
  MMLE}: \ie,
\begin{eqnarray*}
\theta_{(t)} \rightarrow \{ \theta: \partial_\theta l(\theta;\mathbf{v}) = 0 \}, \text{ a.s.}.
\end{eqnarray*}

\medskip
\subsection{Case 2: $\frac{a_{(t)}}{b_{(t)}} \to \infty$}

In \underline{\bf Case 2}, we have following two properties: 
\begin{compactenum}[\bf P1.]
\item For all $T>0$, almost surely,
\begin{eqnarray}
\label{eq:the4}
\lim_{s \to \infty} \sup_{\tau \in [s,s+T]} \| \theta_\alpha(\tau) -
\theta_\alpha(s) \| = 0.
\end{eqnarray}
\item Almost surely, 
\begin{eqnarray*}
\lim_{\tau \to \infty} \| \hat{\mu}_\beta(\tau) -
\hat{\mu}^*(\theta_\beta(\tau);\mathbf{v}) \| = 0.
\end{eqnarray*}
\end{compactenum}


{\bf P1} states that $\theta_\alpha(\tau)$ almost behaves like a
constant after a sufficient number of iterations. This is due to the
fact that $\theta_{(t)}$ is updated by the step-size $b_{(t)}$, but
$\theta_\alpha(\tau)$ is the trajectory made by the slower time-scale of
$a_{(t)}$. More formally, rewriting \eqref{eq:SAP-E}, we have:
\begin{eqnarray*}
  \theta_{(t+1)} &=& \theta_{(t)} + a_{(t)} \cdot \bigg[
  \frac{b_{(t)}}{a_{(t)}} u \bigg( \hat{\mu}_{(t)}, \theta_{(t)},
  X_{(t+1)}  \bigg) \bigg],
\end{eqnarray*}
and thus it is obvious that its limiting ODE is $\dot{\theta}(\tau) =
0$. Then, the property {\bf P1} immediately holds. {\bf P2} implies that
$\hat{\mu}_\beta(\tau)$ is asymptotically close to a unique fixed point
$\hat{\mu}^*(\theta_\beta(\tau); \mathbf{v})$, the expectation of
empirical mean parameter over the distribution $p_{\theta_\beta(\tau)}(h
| \mathbf{v})$ for a given $\theta_\beta(\tau)$ and $\mathbf{v}$. It is
clear that the map $\hat{\mu}^*(\cdot;\mathbf{v})$ is a bijection for a
regular and minimal exponential family. In the rest of the proof, as in
\underline{\bf Case 1}, we first show {\bf P2} in the first step, and
then in the second step, we complete the proof using {\bf P1} and {\bf
  P2}. 

\medskip

{\bf \em Step 1:} \underline{\em Understanding the asymptotic behavior
  of the system at the faster time-scale $\alpha$.}

We introduce $\{\hat{\mu}_\alpha^s(\tau): \tau \in \mathbb{R}_+ \}$,
which interpolates $\{\hat{\mu}_{(t)}^s: t \in \mathbb{Z}_{\geq 0}\}$,
where $\{\hat{\mu}_{(t)}^s: t \in \mathbb{Z}_{\geq 0}\}$ is constructed
such that with $s \in \mathbb{R}$, $\hat{\mu}_{(t)}^s = \hat{\mu}_{(t)}$
for $s \geq \alpha(t)$, and 
\begin{eqnarray} \label{eq:the5}
\hat{\mu}_{(t+1)}^s = \hat{\mu}_{(t)}^s + a_{(t)} \cdot g \bigg(
\hat{\mu}_{(t)}^s, \theta_\alpha(s), H_{(t+1)} \bigg),
\end{eqnarray}
for $s < \alpha(t)$. Note that \eqref{eq:the5} is slightly different to
\eqref{eq:SAP-E} in that $\theta_{(t)}$ is fixed to
$\theta_\alpha(s)$. Then, from the Lipschitz continuity of $g(\cdot)$,
we get that for all $T > 0$:
\begin{eqnarray} \label{eq:the6}
\lim_{s \to \infty} \sup_{\tau \in [s,s+T]} \| \hat{\mu}_\alpha^s(\tau)
- \hat{\mu}_\alpha(\tau) \| = 0.
\end{eqnarray}

Now, we will compare $\hat{\mu}_\alpha^s(\tau)$ to the solution
trajectory of the following ODE, as in Section~\ref{sec:SA-CMN}:
\begin{eqnarray} \label{eq:ODE-E2}
\dot{\mu}(\tau) = g^\ddagger(\mu(\tau)) \defeq \sum_{h \in \mathcal{X}^h}
g \bigg( \mu(\tau), \theta_\alpha(s), \phi(\mathbf{v},h) \bigg) p_{\theta_\alpha(s)}(h|\mathbf{v}).
\end{eqnarray}
To see how the setup matches with that in Section~\ref{sec:SA-CMN}, let
$\tilde{\mu}^s(\tau)$ be the solution on $[s,s+T]$ (for $T>0$) of the
ODE \eqref{eq:ODE-E2} with $\tilde{\mu}^s(s) =
\hat{\mu}_\alpha^s(s)$. Then, $\{ \hat{\mu}_{(t)}^s: t \in
\mathbb{Z}_{\geq 0}\}$ is a discrete-time stochastic process with
controlled Markov process considered in \eqref{eq:SA-cont} for $s \leq
\alpha(t)$. We can verify that the assumptions {\bf (C1)-(C4)} are
satisfied as we verified in the proof for \underline{\bf Case 1}. Then,
we have that for all $T>0$, 
\begin{eqnarray*}
\lim_{s \to \infty} \sup_{\tau \in [s,s+T]} \| \hat{\mu}_\alpha(\tau) -
\tilde{\mu}^s(\tau) \| = 0, \quad \text{ a.s.},
\end{eqnarray*}
which is a direct consequence of Theorem~\ref{thm:SA-CMN} and
\eqref{eq:the6}.

Here, it is clear that the ODE system \eqref{eq:ODE-E2} has a unique
fixed point $\hat{\mu}^* = \hat{\mu}^*(\theta_\alpha(s); \mathbf{v})$,
\ie, the expectation of empirical mean parameter over
$p_{\theta_\alpha(s)}(\cdot | \mathbf{v})$, from the fact that 
\begin{eqnarray*}
g^\ddagger(\mu(\tau)) = \mathbb{E}_{\theta_\alpha(s),\mathbf{v}}[\phi(\mathbf{v},H)] -
\mu(\tau).
\end{eqnarray*}
Then, by applying Lemma~\ref{lem:step1} to this framework we have almost
surely, 
\begin{eqnarray} \label{eq:fast-E}
\| \hat{\mu}_{(t)} - \hat{\mu}^*(\theta_{(t)}; \mathbf{v}) \| \to 0,
\end{eqnarray}
which in turn implies that under the slower time-scale $\beta$, we also
have almost surely, 
\begin{eqnarray} \label{eq:step1-E}
\lim_{\tau \to \infty} \| \hat{\mu}_\beta(\tau) -
\hat{\mu}^*(\theta_\beta(\tau); \mathbf{v}) \| = 0.
\end{eqnarray}
This completes the proof of {\bf P2}.


\medskip

{\bf \em Step 2.} \underline{\em Understanding the asymptotic behavior
  of the system at the slower time-scale $\beta$.}

We start by introducing $\{ \theta_\beta^s(\tau): \tau \in
\mathbb{R}_+\}$, which interpolates $\{\theta_{(t)}^s: t \in
\mathbb{Z}_{\geq 0} \} $, where it is constructed such that with $s \in
\mathbb{R}$, $\theta_{(t)}^s = \theta_{(t)}$ for $s \geq \beta(t)$, and 
\begin{eqnarray} \label{eq:theta1}
\theta_{(t+1)}^s = \theta_{(t)}^s + b_{(t)} \cdot u\bigg(
\hat{\mu}^*(\theta_{(t)}^s;\mathbf{v}), \theta_{(t)}^s, X_{(t+1)}\bigg)
\end{eqnarray}
for $s < \beta(t)$. Note that \eqref{eq:theta1} is slightly different to
\eqref{eq:SAP-M} in that $\hat{\mu}_{(t)}$ is fixed to
$\hat{\mu}^*(\theta_{(t)}^s;\mathbf{v})$. Then, from \eqref{eq:step1-E}
and the Lipschitz continuity of $u(\cdot)$, it follows that for all
$T>0$, 
\begin{eqnarray} \label{eq:theta2}
\lim_{s \to \infty} \sup_{\tau \in [s,s+T]} \| \theta_\beta^s(\tau) -
\theta_\beta(\tau) \| = 0.
\end{eqnarray}

Next, we compare $\theta_\beta^s(\tau)$ to the solution trajectory of
the following ODE:
\begin{eqnarray} \label{eq:ODE-M2}
\dot{\theta}(\tau) = u^\ddagger(\theta(\tau)) \defeq \sum_{x \in
  \mathcal{X}} u \bigg( \hat{\mu}^*(\theta(\tau);\mathbf{v}),
\theta(\tau), \phi(x) \bigg) p_{\theta(\tau)}(x).
\end{eqnarray}
It is clear that the process $\{ \theta_{(t)}^s \}$ is also a
discrete-time stochastic process with controlled Markov process in
Section~\ref{sec:SA-CMN} for $s \le \beta(t)$. The assumptions {\bf
  (C1)-(C4)} are verified similarly in \underline{\bf Case 1}. Let denote
by $\tilde{\theta}^s(\tau)$ be the solution on $[s,s+T]$ (for any $T>0$)
of the ODE \eqref{eq:ODE-M2} with $\tilde{\theta}^s(s) =
\theta_\beta^s(s)$. Then, we have that for all $T>0$,
\begin{eqnarray*}
\lim_{s \to \infty} \sup_{\tau \in [s,s+T]} \| \theta_\beta(\tau) -
\tilde{\theta}^s(\tau) \| = 0, \quad \text{ a.s.},
\end{eqnarray*}
which is a direct result from Theorem~\ref{thm:SA-CMN} and
\eqref{eq:theta2}.


\paragraph{Remark 4.} We can shortly claim that $V(\theta) = -l(\theta;
\hat{\mu}^*(\theta; \mathbf{v}))$ is a Lyapunov function of the ODE
\eqref{eq:ODE-M2}, by checking the definition of Lyapunov function in
\cite{borkar08SA}. Moreover, $l(\theta; \hat{\mu}^*(\theta; \mathbf{v}))
= l(\theta;\mathbf{v})$, which is the exact bound of the marginal
log-likelihood in \eqref{eq:log_lbnd}.

Then, we plug the process $\{ \theta_{(t)}: t \in \mathbb{Z}_{\geq 0}\}$
and ODE \eqref{eq:ODE-M2} into Lemma~\ref{cor:lyap}, \ie, convergence
analysis of stochastic approximation procedure with Lyapunov function,
and we have that:
\begin{eqnarray} \label{eq:M-lyap}
\theta_{(t)} \to \{ \theta: \partial_\theta l(\theta;\mathbf{v}) = 0 \}, \quad
\text{ a.s.}.
\end{eqnarray}
This completes the proof of {\bf \em Step 2}. Combining {\bf \em Step 1}
and {\bf \em Step 2}, \ie, from \eqref{eq:fast-E} \eqref{eq:M-lyap},
completes the proof for \underline{\bf Case 2}.

\medskip
Finally, for both \underline{\bf Case 1} and \underline{\bf Case 2}, we
conclude that under APCD, $\theta_{(t)}$ almost surely converges to a
stationary point of {\bf MMLE}, \ie, 
\begin{eqnarray*}
\theta_{(t)} \rightarrow \{ \theta: \partial_\theta l(\theta;\mathbf{v})
= 0 \}, \quad \text{ a.s.},
\end{eqnarray*}
which completes the proof of Theorem~\ref{thm:EM}. $\qed$

\end{document}